\documentclass[
]{ceurart}

\usepackage{listings}
\usepackage{xspace}
\usepackage{graphicx}
\usepackage{tabularx}
\usepackage{subcaption}
\newcommand{\vago}[0]{\textsc{VAGO}\xspace}

\newcommand{\roberta}[0]{\textsc{RoBERTa}\xspace}
\newcommand{\sBERT}[0]{\textsc{sBERT}\xspace}
\newcommand{\BERT}[0]{\textsc{BERT}\xspace}

\newcommand{\PE}[1]{\textcolor{black}{#1}}

\usepackage{amsmath} 
\usepackage{booktabs}
\usepackage{makecell}

\begin{document}

\copyrightyear{2024}
\copyrightclause{Copyright for this paper by its authors.
  Use permitted under Creative Commons License Attribution 4.0
  International (CC BY 4.0).}

\conference{CLEF 2024: Conference and Labs of the Evaluation Forum, September 09–12, 2024, Grenoble, France}

\title{HYBRINFOX at CheckThat! 2024 - Task 2: Enriching BERT Models with the Expert System VAGO for Subjectivity Detection}
%A Hybrid Method combining BERT models...

%\tnotetext[1]{You can use this document as the template for preparing your
  %publication. We recommend using the latest version of the ceurart style.}

\title[mode=sub]{Notebook for the HYBRINFOX Team at CheckThat! 2024 - Task 2}


\author[1]{Morgane Casanova}[%
email=morgane.casanova@irisa.fr
]
\fnmark[1]
\address[1]{Université de Rennes, CNRS, Inria, IRISA, France}

\author[2]{Julien Chanson}[email=julien.chanson@mondeca.com]
\fnmark[1]
\address[2]{Mondeca, France}

\author[3]{Benjamin Icard}[% 3
orcid=0009-0005-4530-5646,
email=benjamin.icard@lip6.fr
]
\fnmark[1]
\address[3]{LIP6, Sorbonne Université, CNRS, France}
\address[4]{Institut Jean-Nicod, CNRS, ENS-PSL, EHESS, France}

\author[5,6]{Géraud Faye}[%
orcid=0000-0002-2985-5964,
email=geraud.faye@centralesupelec.fr
]
% \cormark[1]

\address[5]{Airbus Defence and Space, France}
\address[6]{Université Paris-Saclay, CentraleSupélec, MICS, France}

\author[5]{Guillaume Gadek}[email=guillaume.gadek@airbus.com]

\author[1]{Guillaume Gravier}
[email=guillaume.gravier@irisa.fr]
\fnmark[1]

\author[4]{Paul \'Egr\'e}[%
orcid=0000-0002-9114-7686,
email=paul.egre@ens.psl.eu
]
\fnmark[1]

%\cortext[1]{Corresponding author.}
\fntext[1]{MC, JC and BI share first authorship on this paper. GGr and PE are co-lead authors.}

\begin{abstract}
This paper presents the HYBRINFOX method used to solve Task 2 of Subjectivity detection of the CLEF 2024 CheckThat! competition. \PE{The specificity of the method is to use a hybrid system, combining a \roberta model, fine-tuned for subjectivity detection, a frozen sentence-\BERT (\sBERT) model to capture semantics, and several scores calculated by the English version of the expert system \vago, developed independently of this task to measure vagueness and subjectivity in texts based on the lexicon. In English, the HYBRINFOX method ranked 1st with a macro F1 score of 0.7442 on the evaluation data.
%Like various methods used in the 2023 edition, the HYBRINFOX method uses a transformer-based approach (based on \BERT), but its distinctive feature is that it enriches it with rule-based scores. %rule-based scores of vagueness, subjectivity, detail, and objectivity for sentences. 
%These scores are calculated using the symbolic system \vago, developed independently of this task to measure vagueness and subjectivity in texts. \vago relies on a lexical database comprised of only English and French at the time of the experiment. 
%In English the HYBRINFOX team \PE{ranked 1st with a macro F1 score of 0.7442} on the test data. %achieved the highest performance. 
For the other languages, the method used a translation step into English, producing more mixed results (ranking 1st in Multilingual and 2nd in Italian over the baseline, but under the baseline in Bulgarian, German, and Arabic).} We explain the principles of our hybrid approach, and outline ways in which the method could be improved for other languages \PE{besides English}. 
\end{abstract}

%The method in fact combines a \roberta model, fine-tuned for subjectivity detection, a frozen sentence-\BERT (\sBERT) model to capture semantics, and the scores calculated by the English version of the expert system \vago

\begin{keywords}
 Subjectivity \sep Objectivity \sep Vagueness \sep Hybrid AI \sep VAGO \sep Large Language Models
\end{keywords}

\maketitle

\section{Introduction}

Detecting subjectivity in natural language is an important task for a
number of applications in the domain of news and communication, with strong ties to
disinformation and propaganda that constitute the playground of the
HYBRINFOX project. % (ANR-21-ASIA-0003-01). 
Indeed, objective statements in news can be defined as statements that are open to verification by limiting bias and interpretive disagreement. 
%typically
%convey verifiable facts. %with little or no bias after verification. 
%As
%such, they are often linked to trustworthy information. 
While an objective statement may turn out false, the information it conveys is generally taken to be trustworthy, because it is prone to independent confirmation 
and fact-checking.
%typically backed up by multiple testimonies.

On the
contrary, subjective statements convey personal feelings and
opinions. By definition, they are prone to inter-personal disagreement and do not obey the same norms of justification and verification. While subjectivity may appear in explicit opinion papers that are not
necessarily considered as propaganda or as manipulation, it is
also widely used implicitly in conjunction with false objective
statements, or just to bias true objective information.

Task 2 of the CLEF 2024 CheckThat!
benchmark~\cite{galassi2023overview,ruggeri2023definition,10.1007/978-3-031-56069-9_62,clef-checkthat:2024:task2}, running for the second
time since 2023, aimed at detecting subjective utterances and thus met
the objectives of the HYBRINFOX project that seek to develop hybrid
methods for the identification of vague information likely to
introduce or encourage bias (subjectivity, evaluativity). In
particular, HYBRINFOX aims to develop tools for measuring linguistic
vagueness in texts, taking advantage of a symbolic AI method, \vago, to
improve the performance of deep learning
models~\cite{guelorget21,icard23}. The project also explores the
boundary between truthful and untruthful uses of linguistic vagueness %and markers of subjectivity
in discourse~\cite{Egre&Icard2018,egre2023optimality}.

Subjectivity, vagueness, uncertainty, speculations, expressions of
opinions are rather difficult to detect automatically in texts as
these concepts involve several pragmatic and rhetorical aspects that go beyond the lexical semantics for which most NLP models are
designed. The difficulty also translates in the definition of
subjectivity, where annotation guidelines have evolved over time --
see,
e.g.,~\cite{wilson2003annotating,kennedy2016two,ruggeri2023definition}. There
is, however, a vast literature on these topics, with a number of recent
contributions as part of previous editions of the CheckThat!
benchmark~\cite{clef-checkthat-23}.

%Subjectivity and vagueness detection has been addressed with expert systems, often relying on lexical analysis as well as on patterns, viz.~\cite{uncertainty,riloff2003learning,icard2022vago}. To some extent, \cite{uncertainty} also makes use of linguistic heuristics to determine the uncertainty scope within an utterance. More recent work, however, relies on statistical models, such as~\cite{huo2020utilizing,sagnika2021attention}, and all systems at the 2023 benchmark were
%based on large language models (LLMs) such as \BERT or
%GPT~\cite{baris2023dwreco,dey2023nn,frick2023fraunhofer,leistra2023thesis,pachov2023gpachov,sadouk2023vrai,tran2023accenture}. Most systems explored fine-tuning different language models for subjectivity detection, either in a multilingual or in a language-specific setting. Some systems added components to cope with data imbalance and scarcity, here again leveraging large language models including generative ones for data augmentation~\cite{baris2023dwreco,frick2023fraunhofer}. While statistical systems do perform well, they lack explicit features of expert systems that make them explainable and that we believe can make them more efficient.

%we believe can be beneficial.

\begin{figure}[tb!]\centering
\includegraphics[scale=.45]{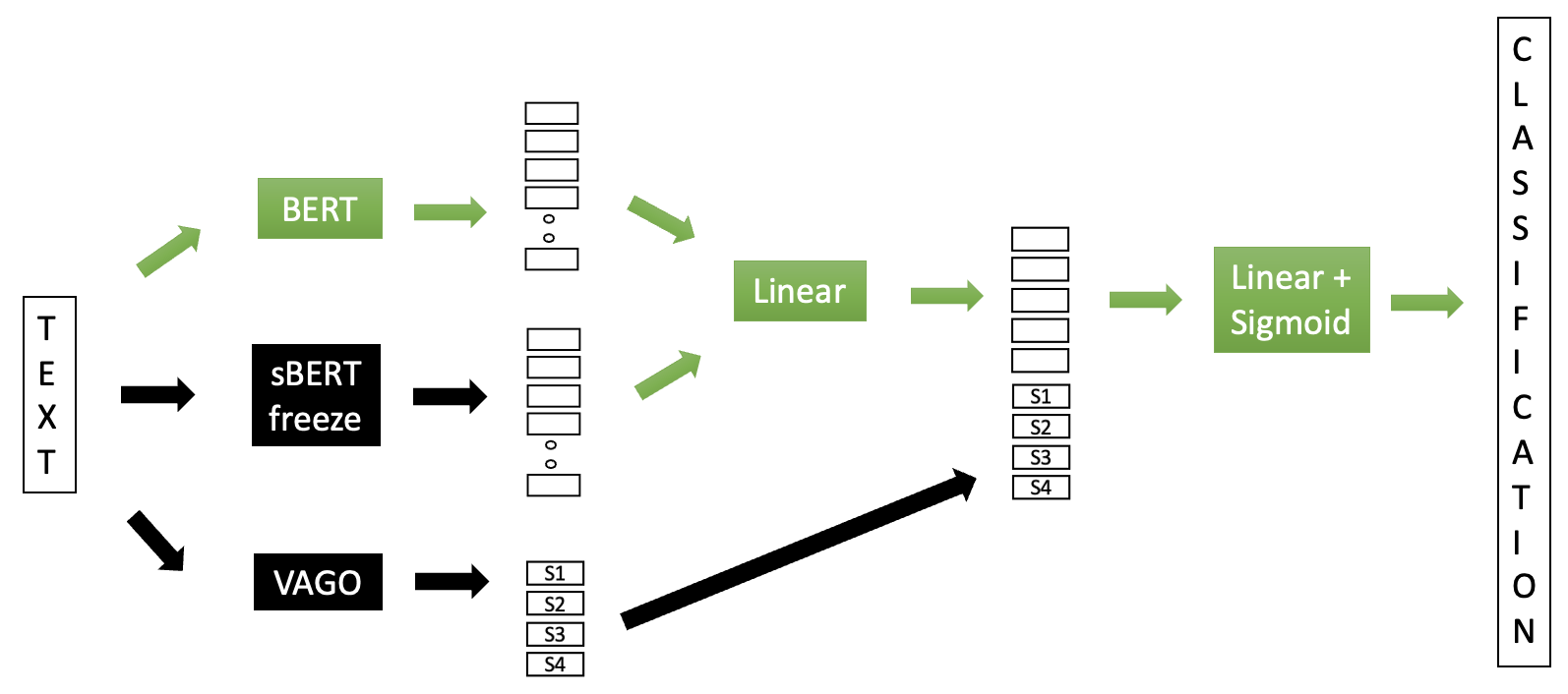}%
\caption{HYBRINFOX model combining \BERT, \sBERT and the four scores (S1, S2, S3, S4) calculated by the expert system \vago for the task of objectivity versus subjectivity classification. The green arrows indicate the elements being trained.}
\label{fig:system}
\end{figure}

The HYBRINFOX method that competed explores the augmentation of an LLM-based system with vagueness scores given by an expert system primarily based on lexical analysis as illustrated in Figure~\ref{fig:system}. The method in fact combines a \roberta model, fine-tuned for subjectivity detection, a frozen sentence-\BERT (\sBERT) model to capture semantics, and the scores calculated by the English version of the expert system \vago \cite{icard2022vago}.\footnote{In Checkthat! 2024 Task 1, our team explored a distinct but similarly hybrid approach for checkworthiness estimation, see \cite{clef-checkthat:2024:task1:hybrinfox} for details.} Dimensionality reduction operates on the concatenation of the \BERT and sentence-\BERT embeddings before combining with the vagueness scores as input to a classification linear layer. \PE{We justify the selection of this package by comparison with other, less efficient combinations, on the developmental data (see Table \ref{tab:bert_performance} below).}

The hybrid system was primarily developed for English, and then adapted to other languages, relying on automatic translation into English for the other test languages. \PE{The method ranked 1st in English on the evaluation data (with a macro F1 score of 0.7442), but in other languages it produced more mixed results (ranking 1st on Multilingual and 2nd in Italian over the baseline, but 3rd in Bulgarian, 4th in German, and 6th in Arabic under the baseline), likely due to loss in accuracy caused by the translation step.}

%In English, the method produced good results, but in other languages the results were much more variable, likely due to loss caused at the translation step.  

%for the expert system part.

The paper is organized as follows. \PE{We start with a brief state of the art in Section~\ref{sec:soa}.} In Section~\ref{sec:vago} we introduce the expert system \vago\ designed to produce vagueness and subjectivity scores. Section~\ref{sec:hybrid} reports on experiments conducted towards the choice of an adequate hybrid system, and presents comparative results on the development set. Section~\ref{sec:results} analyzes the results obtained in the evaluation phase, including post hoc analyses. \PE{Finally, Section~\ref{sec:conclusion} concludes with suggestions on how to extend and improve the current results.}

\section{State of the art}\label{sec:soa}

Subjectivity and vagueness detection has been addressed with expert systems, often relying on lexical analysis as well as on patterns, viz.~\cite{uncertainty,riloff2003learning,icard2022vago}. To some extent, \cite{uncertainty} also makes use of linguistic heuristics to determine the uncertainty scope within an utterance. More recent work, however, relies on statistical models, such as~\cite{huo2020utilizing,sagnika2021attention}, and all systems at the 2023 benchmark were
based on large language models (LLMs) such as \BERT or
GPT~\cite{baris2023dwreco,dey2023nn,frick2023fraunhofer,leistra2023thesis,pachov2023gpachov,sadouk2023vrai,tran2023accenture}. Most systems explored fine-tuning different language models for subjectivity detection, either in a multilingual or in a language-specific setting. Some systems added components to cope with data imbalance and scarcity, here again leveraging large language models including generative ones for data augmentation~\cite{baris2023dwreco,frick2023fraunhofer}. While statistical systems do perform well, they lack explicit features of expert systems that make them explainable and that we believe can make them more efficient. \PE{Conversely, expert systems make very limited use of contextual features. The gist of our approach, therefore, is to define a method drawing on both approaches and combining their respective strengths.}

%%%%%%%%%%%%%%%%%%%%%%%%%%%%%%%%%%%%%%%%%%%%%%%%%%%%%%%%%%%%%%%%%%%%%%%%%%%%%%%%%%%%
%%%%%%%%%%%%%%%%%%%%%%%%%%%%%%%%%%%%%%%%%%%%%%%%%%%%%%%%%%%%%%%%%%%%%%%%%%%%%%%%%%%%
\section{Symbolic scoring using \vago}
\label{sec:vago}

\vago relies on a symbolic approach to assign scores of vagueness, subjectivity, detail, and objectivity to sentences. %\footnote{In \cite{icard&alWI-IAT2023}, precision is called detail, the idea being that named entities provide details on a situation. We use the term ``precision'' here since detail is taken to provide a positive measure of precision.} 
Regarding the measure of detail and objectivity of a text, the detection is based in part on identifying named entities (including people, locations, temporal indications, institutions, and numbers), using the open-source library for Natural Language Processing spaCy.\footnote{\url{https://spacy.io/}} Our underlying assumption is that such entities ground the information reported in specific objects and generally leave very limited room for variable interpretation. The more named entities, the more detailed the sentence is likely to be. Given a sentence $\phi$, we say that its category is $NE$ if it names an entity. The class $NE$ is not closed, as its members are determined by named entity recognition.

By contrast, the detection of vagueness and subjectivity relies on a closed but evolving lexical database, which consisted of 1,614 terms in English at the time of the CheckThat! 2024 experiment \cite{datasetvago2022}. Derived from a typology of vagueness proposed in \cite{Egre&Icard2018}, this database provides an inventory of lexical items distributed in four categories for vagueness (approximation vagueness ($V_A$), generality vagueness ($V_G$), degree vagueness ($V_D$), combinatorial vagueness ($V_C$), and an additional category of explicit markers of subjectivity ($E_S$), not counted as vague.

Expressions of approximation vagueness include modifiers like ``approximately'', which relax the truth conditions of the modified expression. Generality vagueness includes determiners like ``some'' and modifiers such as ``at most''. %Unlike expressions of approximation, the latter have precise truth conditions. 
The category of expressions related to degree vagueness and combinatorial vagueness \cite{alston1964philosophy} mainly consists of one-dimensional gradable adjectives (such as ``tall'' and ``old'') and multidimensional gradable adjectives, including a number of evaluative adjectives (like ``beautiful'', ``intelligent'', ``good'', or ``skilled''). These expressions, unlike expressions of generality vagueness, lack precise truth-conditions, and they leave room for inter-personal disagreement and subjectivity~\cite{wright1995epistemic,kennedy2013two,verheyen2018subjectivity,Solt2018}. Finally, explicit markers of subjectivity include separate expressions such as exclamation marks (``!''), first-person pronouns (``I/we''), and some expressive adverbs (``ever'', ``of course''). \PE{As seen in Table \ref{tab:database}, the category $V_C$ is much bigger than the others, and as we will see below it also plays a determinant role in the detection of subjectivity.%Some of the items in it are less frequent than items in other categories having fewer members.
}

Expressions of type $NE$, $V_A$, $V_G$, are treated as factual or \textit{objective}, whereas expressions of type $E_S$, $V_D$, $V_C$, are treated as \textit{subjective}. Table \ref{tab:database} provides the exact numbers of items per category available in the English \vago database used for the CheckThat! 2024 competition (version of May 2024).

\begin{table}[tb!]
\caption{Number of entries in the English \vago database used in CheckThat! 2024.}
\label{tab:database}
\centering
\begin{tabular}{lcccr}
\toprule
\vago categories  & & Labelling & & Number of entries  \\
\midrule
Approximation & & $V_A$& & 9 \\
Generality & & $V_G$ && 35 \\
Degree vagueness && $V_D$ && 57 \\
Combinatorial vagueness && $V_C$ && 1,500 \\
Explicit subjectivity  && $E_S$ && 13 \\
\midrule
All categories && && 1,614 \\
\bottomrule
\end{tabular}\medskip

\end{table}

For each measurement, relevant markers are detected and scored from words to sentences. For a given sentence $\phi$, its \textit{vagueness score} is calculated as the ratio between the sum of vague words in the sentence, $|V|_{\phi}$, and the total number of words in the sentence, notated $N_{\phi}$. That is:
\begin{align}
\label{eq: vague}
R_{vagueness}(\phi) =
\frac{\overbrace{|V_D|_{\phi} + |V_C|_{\phi}}^{\text{subjective}} + \overbrace{|V_A|_{\phi} + |V_G|_{\phi}}^{\text{objective}}}{N_{\phi}}
= \frac{|V|_{\phi}}{N_{\phi}}
\end{align}

\noindent where $|V_A|_{\phi}$, $|V_G|_{\phi}$, $|V_D|_{\phi}$, and $|V_C|_{\phi}$ represent the number of terms in $\phi$ belonging to each of the four vagueness categories. %(approximation, generality, degree vagueness, and combinatorial vagueness).

The \textit{subjectivity score} of a sentence $\phi$ is calculated as the ratio between the subjective expressions in $\phi$, either vague or explicit markers, and the total number of words in $\phi$. \PE{That is, letting $|S|_{\phi}=|E_S|_{\phi}+|V_D|_{\phi} + |V_C|_{\phi}$:} 
\begin{align}
R_{subjectivity}(\phi) &= \frac{|S|_{\phi}}{N_{\phi}}
\end{align}

The \textit{detail-vs-vagueness score} of a sentence $\phi$ is defined as the relative proportion of named entities $|NE|_{\phi}$ in the sentence, compared to the number of vague terms in $\phi$ (across all categories) $|V|_{\phi}$:
\begin{align}
R_{detail/vagueness}(\phi) &=
\frac{|NE|_{\phi}}{|NE|_{\phi} + |V|_{\phi}}\label{eq:precision}
\end{align}

The \textit{objectivity-vs-subjectivity score} of a sentence is defined as the relative proportion of objective expressions in $\phi$ (objective vagueness and named entities), compared to the number of subjective terms in $\phi$. \PE{That is, letting  $|O|_{\phi} = |NE|_{\phi} + |V_A|_{\phi} + |V_G|_{\phi}$:}
\begin{align}
R_{objectivity/subjectivity}(\phi) &=
\frac{|O|_{\phi}}{|O|_{\phi}+|S|_{\phi}}\label{eq:os}
\end{align}

%\noindent where $|O|_{\phi}$ is the aggregate number of expressions of objectivity in $\phi$.

%, namely $|O|_{\phi} = |NE|_{\phi} + |V_A|_{\phi} + |V_G|_{\phi}$; and where $|S|_{\phi}$ is the aggregate number of expressions of subjectivity in $\phi$, namely $|S|_{\phi} = |E_S|_{\phi}+|V_D|_{\phi} + |V_C|_{\phi}$.\\

In summary, the symbolic method encodes objectivity in terms of two main dimensions: named entities ($NE$) and objective vague expressions. And likewise it encodes subjectivity in terms of subjective vague expressions and explicit markers of subjectivity. Hence, vagueness and subjectivity overlap, but neither includes the other, and vagueness does not rule out objectivity. Expressions of detail are all markers of objectivity, but objectivity is a broader category.%\medskip

\vago does not consist solely of a lexicon and scoring rules, but it also includes expert rules of vagueness-cancellation (viz. the measure phrase ``180cm'' in ``Mary is 180cm tall'' cancels the vagueness and subjectivity of the adjective ``tall'' \PE{when it occurs unmodified as in ``Mary is tall''}). Quotation marks are also handled as cancelling the subjectivity of terms occurring within the marks. This choice agrees with with the annotation guide proposed in \cite{ruggeri2023definition}, in which reported speech is taken to be conveying objective information on views that may themselves be subjective.
%TOBEADDED: mention the management of quote signs, and ways in which \vago takes into account some aspects of the annotation guide for subjectivity.

%%%%%%%%%%%%%%%%%%%%%%%%%%%%%%%%%%%%%%%%%%%%%%%%%%%%%%%%%%%%%%%%%%%%%%%%%%%%%%%%%%%%
%%%%%%%%%%%%%%%%%%%%%%%%%%%%%%%%%%%%%%%%%%%%%%%%%%%%%%%%%%%%%%%%%%%%%%%%%%%%%%%%%%%%
\section{Development phase: defining an optimal hybrid system}
%Hybrid classification}
\label{sec:hybrid}

During the development phase, we tested and compared six variants of our system on the English data. Two variants are purely machine-learning based and serve as a baseline to measure the contribution of \vago to a hybrid system. The other four variants address different ways of integrating \vago features to a machine-learning approach. Each system was fine-tuned on the English training data over 30 epochs, with a batch size of 6 and a learning rate of 10e-6. Results for the different systems are reported in Table~\ref{tab:bert_performance}. For languages other than English, we used the English system on an initial translation into English using the DeepL translator\footnote{\url{https://www.deepl.com/}}. We thus only report results on English for the development phase.

\begin{table}[tb!]
\caption{Performance of different \BERT-based systems tested on the development data (English).} %The optimal system is marked with a triangle.}
    \label{tab:bert_performance}
    \centering
    \begin{tabular}{l c c c c}
        \toprule
        \multicolumn{1}{l}{\textbf{Systems}} & \textbf{Threshold} & \textbf{Macro F1} & \textbf{SUBJ F1} & \\
        \midrule
        \makecell[l]{\roberta} & 0.1 & 0.786 & 0.7851 & \\
                      
          \hline

        \makecell[l]{\roberta + \sBERT} & 0.15 & 0.8025 & 0.8033 & \\
        %\addlinespace
                \hline
        \makecell[l]{\roberta + \vago Terms} & 0.3 & 0.802 & 0.8108 & \\
        %\addlinespace
                        \hline
        \makecell[l]{\roberta + \vago Scores} & 0.05 & 0.817 & 0.8258 & \\
                \hline
        %\addlinespace
        \makecell[l]{\roberta + \sBERT + \vago Scores} & 0.1 & 0.8173 & 0.8346 & $\triangleleft$ \\
                \hline
        %\addlinespace
        \makecell[l]{\roberta + \sBERT + \vago Terms + \vago Scores} & 0.2 & 0.8144 & 0.8099 & \\ 
        \bottomrule
    \end{tabular}
\end{table}

As an initial baseline, a first pure machine-learning system consists in fine-tuning a \BERT model for document classification,  specifically the ``\roberta-base'' model, with a single value output 1 for subjective, and 0 for objective. \PE{We chose \roberta\ based on its good performances on text classification and for reasons of familiarity, leaving open whether other models could be used instead.}

\begin{figure}[tb!]\centering
\includegraphics[scale=.57]{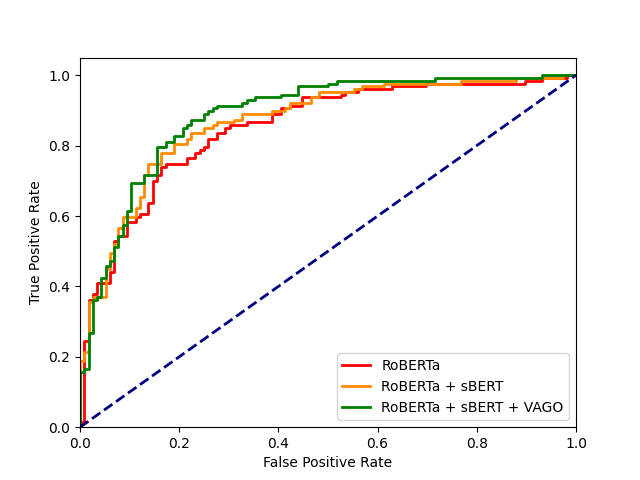}%
\caption{Receiver Operating Characteristic (ROC) Curve for the ``\roberta'', the ``\roberta + \sBERT'' and the ``\roberta + \sBERT + \vago Scores'' Systems Predictions.}
\label{fig:ROC}
\end{figure}

The optimization criterion at training is the binary cross-entropy. At test time, we compare the score given by the system to a threshold, utterances with a score above the threshold being deemed as subjective.\footnote{Note that training is thus done with an implicit threshold equal to 0.5.} By varying the threshold, we obtain different trade-offs in terms of misses and false alarms for subjective utterances, yielding ROC curves as reported in Figure~\ref{fig:ROC}, where the red curve corresponds to the baseline. We defined the optimal threshold on the test set of the development data, searching for the threshold maximizing the official macro F1 metric. Optimal thresholds found on the validation data along with the corresponding scores are reported in Table~\ref{tab:bert_performance}, where the $\triangleleft$ symbol shows the system retained.

Due to the fine-tuning of all the parameters, the resulting model is likely to encode mostly lexical information dedicated to the task at hand, disregarding semantics. We thus tested a variant combining the \BERT model with a sentence-\BERT model whose parameters are frozen. The sentence-\BERT model is a \BERT model trained within a siamese architecture to yield sentence embeddings that are close one to another in the embedded space for utterances having similar meanings. These models, trained on paraphrasing and on natural language inference tasks, are known to encode semantics to some extent, thus providing our system with an intuition of the meaning to facilitate classification. In the experiments, we used the ``distilbert-base-nli-mean-tokens'' checkpoint and did not re-estimate the parameters at training time to keep general semantics. Adding a semantic description as input to the classifier improved results as reported in Table~\ref{tab:bert_performance}.

Making use of the expert system \vago in a \BERT-based approach can rely on one of %one of 
two features: (i)~the terms detected by \vago for the five categories ($V_A, V_G, V_D,V_C, E_S$) in an utterance, referred to as ``\vago Terms''; (ii)~the four vagueness scores described in the previous section, referred to as ``\vago Scores''.

In the first case, a straightforward approach simply consists in augmenting the input utterance with the \vago terms before training a \BERT classifier. The input to the \BERT classifier thus consists of the utterance to classify, followed by a list of \vago terms separated from the utterance with a [SEP] token. The idea of this approach, designated as ``\roberta+ \vago Terms'', is to reinforce the decision by emphasizing the terms deemed of interest by \vago. In the case of \vago scores, the idea is to combine the \BERT utterance embedding with the \vago scores before the classification. To prevent the four \vago scores from being overwhelmed by the 768 dimensions of the \BERT utterance embedding, we first reduced the dimension of the latter to 5 before concatenation with the \vago scores. The resulting 9 dimensional feature vector constitutes the input to the classification head. Both strategies improve performance over the \BERT baseline, with \vago scores being slightly more efficient than \vago terms.

We also combined these last two hybrid approaches with the sentence-\BERT embeddings as discussed previously. The combination of \BERT and sentence-\BERT embeddings with the \vago Scores defines the official HYBRINFOX system that competed in CLEF, achieving the best results on the development set. It is illustrated in Figure~\ref{fig:system} and marked with the symbol $\triangleleft$ in Table~\ref{tab:bert_performance}. The \BERT and sentence-\BERT embeddings are concatenated before reducing the 2x768 dimensional vector to 5 with a linear projection. As previously, this last feature vector is concatenated with the \vago Scores before entering classification layer. We also tested the addition of the \vago Terms to the \BERT encoder in this last architecture, however with no success, yielding a reduction of the subjective F1 score. Interestingly, the ROC curves show a strong benefit for the ``\roberta + \sBERT+ \vago Scores'' over the baseline, where a significant improvement of the false positive rate of 0.25 is observed for a fixed true positive rate of 90\,\%.

%%%%%%%%%%%%%%%%%%%%%%%%%%%%%%%%%%%%%%%%%%%%%%%%%%%%%%%%%%%%%%%%%%%%%%%%%%%%%%%%%%%%
%%%%%%%%%%%%%%%%%%%%%%%%%%%%%%%%%%%%%%%%%%%%%%%%%%%%%%%%%%%%%%%%%%%%%%%%%%%%%%%%%%%%
\section{Evaluation phase: results and analyses}\label{sec:results}

Table~\ref{tab:results-eval} presents the evaluation results with the selected ``\roberta + \sBERT+ \vago Scores'' system, namely the performance of our method on the evaluation data provided at the time of the contest. \PE{Overall, the results were competitive and outperformed the baseline used by the organizers (a logistic regression model) in half of the cases.} 

\PE{In English, our main target and our pivot language, it came out first (out of 15 teams, with a macro-F1 of 0.7442 for Hybrinfox vs 0.6346 for the baseline).} \PE{It also obtained good scores in Italian (0.7838 vs 0.6503, rank=2/5), and in the Multilingual task (0.6849 vs 0.6697, rank=1/3). The scores were less good and under the baseline in Bulgarian (0.7147 vs 0.7531, rank=3/5), German (0.6968 vs 0.6994, rank=4/4), and Arabic (0.4551 vs 0.4908, rank=6/7).} % The largest gap was in German, where the 1st team (nullpointer) obtained a macro F1 of 0.7908.} %The English system, which corresponds to the pivot language of our system, ranked 1st in the evaluation. 

Notably, results in Arabic were much lower than for other languages, though our scores were of the same order of magnitude as those of other participants. This suggests that the Arabic dataset presents specific properties compared to others, making them worthy of delving further into the data and annotations.

We also conducted a post-evaluation analysis of the optimal decision threshold, which was empirically set at 0.1 on the development data. Results are reported in the right-most columns of Table~\ref{tab:results-eval}. Overall, the threshold of 0.1 appears to be stable across languages, however it is suboptimal for German, Arabic and for the Multilingual dataset. For these three datasets, better performance would have been achieved by lowering the threshold to 0.05. In other words, with a threshold of 0.1, the miss rate for subjective utterances was probably too high to yield a good compromise between recall and precision for an optimal macro F1 score. This observation calls for further work on score normalization towards better stability across languages and datasets.

\begin{table}[tb!]
\caption{Results of the ``\roberta + \sBERT + \vago Scores'' system on the evaluation phase datasets. Left column: settings at the time of CheckThat!2024; Right column: optimal settings calculated post hoc.}
\label{tab:results-eval}
\centering
%\begin{tabular}{lccc}
\begin{tabular}{l|ccc||ccc}
\toprule
\textbf{Language} & \textbf{Threshold} & \textbf{Macro F1} & \textbf{SUBJ F1} & \makecell{\textbf{Optimal} \textbf{Threshold}} & \textbf{Macro F1} & \textbf{SUBJ F1} \\
\midrule
English & 0.1 & 0.7442 & 0.6 & 0.1 & 0.7442 & 0.6\\
Italian & 0.1 & 0.7838 & 0.68 & 0.1 & 0.7838 & 0.68\\
Arabic & 0.1 & 0.4551 & 0.27 & 0.05 & 0.4924 & 0.39\\
Bulgarian & 0.1 & 0.7147 & 0.65 & 0.1 & 0.7147 & 0.65\\
German & 0.1 & 0.6968 & 0.57 & 0.05 & 0.7708 & 0.69\\
Multilingual & 0.1 & 0.6849 & 0.63 & 0.05 & 0.7100 & 0.70\\
\bottomrule
\end{tabular}\medskip

\end{table}

% À mettre plutôt dans les discussions  ?

Upon further analysis, we discovered that our results for Bulgarian could have been improved through additional corpus cleaning. Specifically, square bracket symbols were inadvertently included in the translations into English, which we did not address during the evaluation phase. By removing these brackets, our results for Bulgarian improved significantly, achieving a Macro F1 score of 0.7561 and a SUBJ F1 score of 0.7122 with an optimal threshold of 0.1. This issue with square brackets was also present in other languages, but did not affect the results as significantly as it did for Bulgarian.

%\begin{figure}[H]\centering
%\includegraphics[scale=.5]%{Figures/F1_thresholds_all_languages.png}%
%\caption{Macro F1 scores across various classification% thresholds for all languages in the evaluation phase.}
%\label{fig:F1_thresholds}
%\end{figure}

%%%%%%%%%%%%%%%%%%%%%%%%%%%%%%%%%%%%%%%%%%%%%%%%%%%%%%%%%%%%%%%%%%%%%%%%%%%%%%%%%%%%
%%%%%%%%%%%%%%%%%%%%%%%%%%%%%%%%%%%%%%%%%%%%%%%%%%%%%%%%%%%%%%%%%%%%%%%%%%%%%%%%%%%%
\section{Conclusion and future work}\label{sec:conclusion}

To solve Task 2 of Subjectivity Detection, we used a hybrid approach combining the rigid symbolic AI system \vago rules with flexible \BERT predictions taking context into account. Our main target was English, one of the languages of \vago, for which we obtained good results in the development and in the evaluation phases. For other languages, we used the English system on automatic translations obtained by DeepL. %, \PE{but with a wider gap between the results of the two phases}. % maybe say more about this before

In order to improve our approach, we believe that separate \vago lexicons should be developed for additional languages besides English and French, in order to get rid of the translation step into English, and to be able to use appropriate \BERT and \sBERT models for each language. The development of an Italian lexicon was under way at the time of the evaluation, but not sufficiently advanced yet to produce satisfactory results. Alternatively, we could aim for a better control of the quality of the translation into English, our assumption being that the lexicon of objectivity and subjectivity ought to be preserved under translation.

Finally, we stress that our approach of subjectivity detection was developed independently of the task. During the training phase, we noticed that some expressions that \vago classifies as subjective, like the vague determiner ``many'', were not systematically associated with the label ``subjective'' (31 out of 42 sentences including ``many'' are labelled as objective in the development set provided ahead of the evaluation phase). We did not try to modify \vago 's classification principles on this or other specific entries. Instead, we left it in place in order to see better if our understanding of subjectivity and the understanding of subjectivity proposed in \cite{ruggeri2023definition} would agree. We were pleased to see that they mostly do, because this lends support to the hypothesis that the expression of subjectivity is characterized in part by the use of a specific lexicon and by specific rhetorical markers.

% - separate \vago lexicons for more languages besides English and French

% \noindent - get rid of translation step into English, and use a separate \BERT for each language

%\noindent - enrich \vago's category $P$ with also pronouns anaphoric with named entities ?

%\noindent - mention the fact that our understanding of subjectivity differs slightly from that of the annotation guide (case of "many")

%%%%%%%%%%%%%%%%%%%%%%%%%%%%%%%%%%%%%%%%%%%%%%%%%%%%%%%%%%%%%%%%%%%%%%%%%%%%%%%%%%%%
%%%%%%%%%%%%%%%%%%%%%%%%%%%%%%%%%%%%%%%%%%%%%%%%%%%%%%%%%%%%%%%%%%%%%%%%%%%%%%%%%%%%
\section*{Acknowledgements}

\PE{We thank two anonymous referees for helpful comments.} This work was supported by the programs HYBRINFOX (ANR-21-ASIA-0003), FRONTCOG (ANR-17-EURE-0017), and THEMIS (n°DOS\-0222794/00 and n° DO\-S0\-222795/00). PE thanks Monash University for hosting him during the writing of this paper, in the context of the program {PLEXUS (Marie Sk\l odowska-Curie Action, Horizon Europe Research and Innovation Programme, grant n°10\-1086295).

% \section*{Declaration of contribution}


\bibliography{biblio}

\end{document}